\documentclass{article}

\usepackage{fullpage, times}

\usepackage[utf8]{inputenc}  
\usepackage[T1]{fontenc}    
\usepackage{url}           
\usepackage{booktabs}     
\usepackage{amsfonts,bm}     
\usepackage{nicefrac}      
\usepackage{microtype}       
\usepackage{tikz}
\usetikzlibrary{fit,positioning,calc,shapes,arrows.meta}
\usepackage{subcaption}

\usepackage{natbib,hyperref}

\usepackage{color,xcolor}
\definecolor{darkblue}{rgb}{0, 0, 0.5}
\definecolor{darkgreen}{rgb}{0, 0.6, 0}
\definecolor{darkred}{rgb}{0.6, 0, 0}
\hypersetup{colorlinks=true, citecolor=darkblue, linkcolor=darkblue, urlcolor=darkblue}

\usepackage{natbib}
\usepackage[colorinlistoftodos]{todonotes} 

\usepackage{amsmath}
\usepackage{amssymb}
\usepackage{mathtools} 
\usepackage{amsthm}

\usepackage{amssymb}
\usepackage{xcolor}
\usepackage{wrapfig}

\usepackage{enumitem}

\setlist{nosep,leftmargin=0.3in} 
\usepackage{multirow}
\usepackage[noend]{algpseudocode}
\usepackage{algorithm}

\usepackage{graphicx}

\usepackage[capitalize,noabbrev,nameinlink]{cleveref}
\newcommand{\light}{ \mathsf{Fetch}}

\usepackage{mdframed}
\mdfsetup{
linecolor=black!40,
linewidth=0.5pt,
innerleftmargin = 2pt,
innertopmargin = 4pt,
innerrightmargin= 2pt,
innerbottommargin = 4pt,
}

\usepackage{tcolorbox}

 \newtcolorbox[crefname={note}{notes}]{notebox}[1][]{%
    colback=blue!5!white,
    colframe=blue!10!white,
    sharp corners,
    before upper={{\bfseries Note~\thetcbcounter}.\ },
    #1
}
\newtcolorbox[auto counter, number within=section,crefname={insight}{insights}]{insightbox}[1][]{%
    colback=blue!5!white,
    colframe=blue!10!white,
    sharp corners,
    before upper={{\bfseries Insight~\thetcbcounter}.\ },
    #1
}

\newtcolorbox[auto counter, number within=section,crefname={insight}{insights}]{recommendationbox}[1][]{%
    colback=green!5!white,
    colframe=green!15!white,
    sharp corners,
    before upper={{\bfseries Recommendation~\thetcbcounter}.\ },
    #1
}

\theoremstyle{plain}
\newtheorem{theorem}{Theorem}[section]

\theoremstyle{definition}

\theoremstyle{remark}

\newlength\savewidth
\usepackage{xcolor}
\usepackage{tabulary}
\newcolumntype{x}[1]{>{\centering\arraybackslash}p{#1pt}}

\title{Efficient Joint Prediction of Multiple Future Tokens }

\author{ Kwangjun Ahn \and  Alex Lamb \and John Langford
}

\date{Microsoft Research\\[1em]
\today}

\begin{document}

\maketitle 

\begin{abstract}
In this short report, we introduce joint multi-token prediction (JTP), a lightweight modification of standard next-token prediction designed to enrich hidden state representations by jointly predicting multiple future tokens.
Unlike previous multi-token prediction approaches, JTP strategically employs teacher forcing of future-tokens through a carefully designed representation bottleneck, allowing the model to encode rich predictive information with minimal computational overhead during training. We show that the JTP approach achieves a short-horizon belief state representation, while popular alternatives for multi-token prediction fail to do so. We demonstrate the effectiveness of our method on the synthetic star graph navigation task from \cite{bachmann2024pitfalls}, highlighting a significant performance improvement over existing methods. This manuscript presents promising preliminary results intended to stimulate further research.


\end{abstract}

\section{Introduction}
Standard large language models are pretrained using the next-token prediction  objective, which, by the chain rule, allows them to model any distribution over sequences:
$\Pr(x_0,...,x_T)=\Pi_{t=0}^T \Pr(x_t \mid x_0,...,x_{t-1})$. However, in practice, this theoretical expressivity is not always realized due to limitations in the model’s representational capacity or the difficulty of reaching an optimal solution through gradient-based optimization \citep{bachmann2024pitfalls, abbe2024fartransformersreasonglobality, hu2024learning}.

In this short report, we  propose  {\bf joint multi-token prediction} (JTP), an exceptionally efficient solution to the limitations of next-token prediction. JTP aims to enrich the hidden representation of the model by predicting the \emph{joint} distribution of  multiple future tokens over a sliding window of size $D$. Our proposed scheme carefully controls the information flow to effectively enrich the hidden representation---unlike existing multi-token prediction schemes \citep{gloeckle2024better,deepseekai2024v3}. This design enables model representations to converge on solutions that are otherwise inaccessible. Notably, JTP introduces only a minimal additional component to the transformer architecture, resulting in negligible overhead while enabling the model to solve problems that conventional next-token prediction cannot.

Importantly, this manuscript aims to share a promising work-in-progress idea with the research community to encourage further exploration. At this stage, we primarily evaluate our approach on the synthetic star graph navigation task from \cite{bachmann2024pitfalls}, a simple task that is known to fail standard next-token predictors. Our experimental results highlight a striking performance gap between our method and existing multi-token prediction approaches.

In \autoref{sec:mtp}, we discuss how JTP works in more depth.  In \autoref{sec:exp-star}, we then experiment with the approach showing it can solve small problems which simple next token prediction, and alternative methods of multitoken prediction cannot. In \autoref{sec:why}, we provide some theoretical discussion covering the computational cost and capabilities of JTP.  We close with \autoref{sec:exp-llm} showing that optimizing JTP and the next token objective are largely compatible on text datasets.

\section{Efficiently Predicting Joint Distributions of Future Tokens}
\label{sec:mtp}

\begin{figure}[h]
\centering
\includegraphics[width=0.85\linewidth]{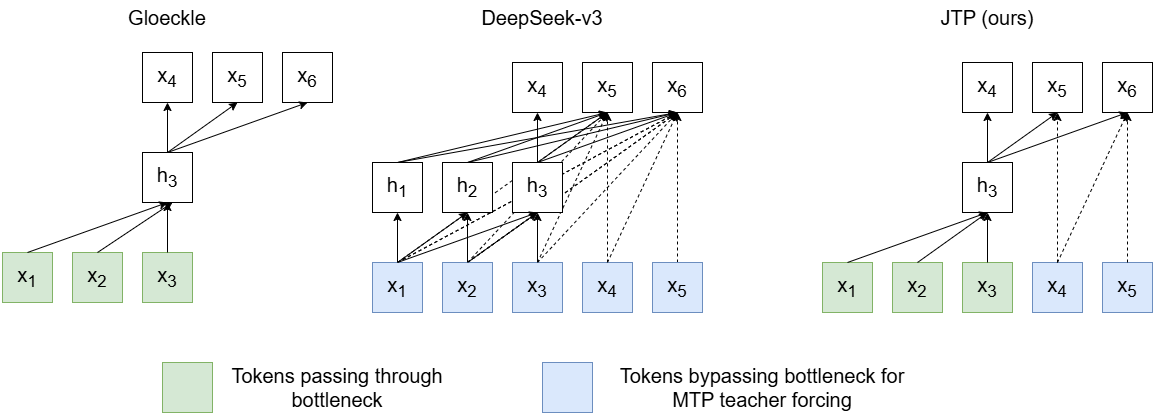}
\caption{Illustration of multi-token prediction mechanisms at position $t=3$. The method of \cite{gloeckle2024better} (left) independently predicts future tokens through a single representation bottleneck, neglecting  dependencies between the future tokens. DeepSeek-V3 \citep{deepseekai2024v3} (middle) processes each token prediction through multiple layers considering the entire historical context, bypassing the desired bottleneck and thus diminishing representation enrichment.   Dotted arrows indicate teacher-forcing dependencies. In contrast, our proposed method (right) efficiently funnels predictive information through a single representation bottleneck while utilizing teacher-forced tokens (dotted arrows), thus preserving token dependencies without compromising representation richness or computational efficiency.}
\label{fig:mtp_comparison}
\end{figure}

In this section, we first introduce our main method in more generality in \autoref{sec:jtp}, and compare our approach with existing multi-token prediction methods in \autoref{sec:comparison}.
We then present an instantiation of our method in \autoref{sec:example}.

Let $\mathbf{h}_{0:T-1}$ denote the hidden states produced by the main Transformer model. Typically, these hidden states are passed through the output head for next-token prediction. The standard next-token prediction loss is given by:
\begin{align*}
\text{Next-Token Loss} ~=~ 
\frac{1}{T}\sum_{t=1}^{T}  \mathcal{L}_{\mathrm{NTP}}(x_t \mid \mathbf{h}_{t-1}) \,,
\end{align*}
where each term is defined as:
\begin{align*}
\mathcal{L}_{\mathrm{NTP}}(x_t \mid \mathbf{h}_{t-1}) ~:=~ - \log \mathrm{head} \bigl(\mathbf{h}_{t-1}\bigr)[x_t].
\end{align*}
Here, $\mathrm{head}$ processes the hidden state $\mathbf{h}_{t-1}$ and outputs the logits for predicting the token $x_t$.

Our goal is to enrich the hidden states $\mathbf{h}_{0:T-1}$ by predicting not only the next token $x_t$ but also several future tokens $x_{t+1}, \ldots, x_{t+D}$. To achieve this, we incorporate a multi-token prediction (MTP) component alongside the standard next-token prediction (NTP) objective:
\begin{align*}
\text{Training Loss} ~=~ \text{Next-Token Loss} ~+~
\lambda\cdot\underbrace{\frac{1}{T}\sum_{t=1}^{T}  \mathcal{L}_{\mathrm{MTP}}(x_{t+1:t+D} \mid \mathbf{h}_{t-1})}_{\text{Multi-Token Loss}} \,.
\end{align*}
Here, $\lambda>0$ balances immediate and future predictions.

Next, we detail the design of the multi-token prediction loss.

\subsection{Joint Multi-Token Prediction (JTP)}
\label{sec:jtp}

Ideally, $\mathcal{L}_{\mathrm{MTP}}$ models the \emph{joint distribution} of the $D$ future tokens $(x_t, x_{t+1}, \ldots, x_{t+D})$ given $\mathbf{h}_{t-1}$. This contrasts with the method of Gloeckle et al.~\cite{gloeckle2024better}, where each future token’s \emph{marginal} distribution is predicted independently (see \autoref{sec:comparison} for details). However, naive joint modeling of $(x_t,x_{t+1}, \ldots, x_{t+D})$ is excessively expensive because the support grows on the order of $\lvert V\rvert^D$.
 
To address this complexity, we apply a teacher-forcing strategy that breaks down the joint distribution via the chain rule. Specifically, we factor $\mathcal{L}_{\mathrm{MTP}}(x_{t+1:t+D} \mid \mathbf{h}_{t-1})$ into a sum of conditionals, each of which can leverage teacher-forced tokens:
\begin{align}
\mathcal{L}_{\mathrm{MTP}}(x_{t+1:t+D} \mid \mathbf{h}_{t-1})
\;=\;
\frac{1}{D}\sum_{i=1}^D \mathcal{L}^i_{\mathrm{MTP}}\bigl(x_{t+i} \,\mid\, \mathbf{h}_{t-1},\, x_{t:t+i-1}\bigr).
\end{align}
A naive implementation of teacher-forcing, however, risks placing too much emphasis on the teacher-forced tokens $x_{t:t+i-1}$. This undermines our goal of enriching the hidden state $\mathbf{h}_{t-1}$ with multi-step “planning” information, since the model could simply rely on those forced tokens and fall back to the next-token prediction mode.

To encourage $\mathbf{h}_{t-1}$ to capture future information, we introduce a lightweight processing module. Instead of passing teacher-forced tokens directly into the main Transformer, we process them through a \emph{light} component, denoted $\light$. This module refines the hidden state $\mathbf{h}_{t-1}$ using the teacher-forced tokens $x_{t:t+i-1}$ to extract relevant information for predicting $x_{t+i}$:
\begin{align*}
    \light_{x_{t:t+i-1}}(\mathbf{h}_{t-1}) \quad \text{extracts information about } x_{t+i}.
\end{align*}

The output of $\light$ is then fed into an MTP head, $\mathrm{head}_{\mathrm{MTP}}$, which performs the prediction. Formally,
\begin{align*}
\mathcal{L}^i_{\mathrm{MTP}}\bigl(x_{t+i} \,\mid\, \mathbf{h}_{t-1},\, x_{t:t+i-1}\bigr)
\;:=\;
-\log \mathrm{head} \Bigl(\light_{ x_{t:t+i-1}}\bigl(\mathbf{h}_{t-1}\bigr)\Bigr)[x_{t+i}].
\end{align*} 
By structuring the model this way, the hidden state $\mathbf{h}_{t-1}$ remains the primary carrier of future-planning information, leading to richer representations. We refer to this approach as \textbf{joint multi-token prediction} (\textbf{JTP}).

We discuss a concrete instantiation of our proposal in \autoref{sec:example} after first highlighting how our proposal differs from existing multi-token prediction schemes next.

\subsection{Comparison with Existing MTPs}
\label{sec:comparison}

Notable existing approaches for multi-token prediction include the frameworks proposed in  \cite{gloeckle2024better} and  \cite{deepseekai2024v3}. Each of these exhibits shortcomings for the purpose of representation enrichment.  (They of course have other virtues.)

\vspace{0.5em}
\noindent\textbf{\cite{gloeckle2024better}.}
Their method independently predicts each of the next \(D\) tokens from the same hidden state, capturing only the \emph{marginal} distribution of each token rather than a coherent joint distribution. Because modeling marginals can require strictly less information than modeling the full joint distribution, the hidden state need not encode all the multi-token dependencies. This undermines the goal of forcing a richer hidden representation.

For instance, consider a short sequence
 $(x_1,x_2,x_3,x_4,x_5,x_6,x_7)$
where $x_1$, $x_2$, $x_3$, $x_4$ are sampled uniformly between $0$ and $1$ independently.  Then define:
\[
(x_5,x_6,x_7) \;=\;
\begin{cases}
(\,x_1,\,x_2,\,x_3) & \text{if } x_4 = 1,\\
(\,1 - x_1,\;1 - x_2,\;1 - x_3) & \text{if } x_4 = 0.
\end{cases}
\]
While the marginal distribution of each future token remains uniform, correctly predicting them \emph{jointly} requires encoding \((x_1,x_2,x_3)\) in the hidden state.  The marginal MTP approach can ignore these dependencies and still match marginals, thus failing to learn a truly enriched state.

\vspace{0.5em}
\noindent\textbf{\cite{deepseekai2024v3}.}
Their MTP method processes the historical context repeatedly with increasing depth to predict future tokens. Specifically, given initial hidden states $\mathbf{h}^0_{0:T-1}=\mathbf{h}_{0:T-1}$ from the main Transformer, their method iteratively applies additional Transformer layers, each incorporating teacher-forced future tokens:
\begin{align}
\mathbf{h}_{0:T-k-1}^{k} &= \mathrm{Transformer}_k\left(  \begin{bmatrix} \mathbf{h}_0^{k-1}\\ x_{k} \end{bmatrix}, \begin{bmatrix} \mathbf{h}_1^{k-1}\\ x_{k+1} \end{bmatrix}, \dots, \begin{bmatrix} \mathbf{h}_{T-k-1}^{k-1}\\ x_{T-k-1} \end{bmatrix}  \right).
\end{align}
Each subsequent Transformer layer thus reprocesses previously obtained representations alongside teacher-forced future tokens. Although this explicitly captures inter-token dependencies, it undermines an intended bottleneck effect, as future token predictions depend on the entire historical sequence of hidden states  and input tokens. 
As a result, the model can rely on the extra layers/contexts at prediction time rather than encoding everything in \(\mathbf{h}_{t}\).
Consequently, this dilutes the enrichment of the primary representation. Additionally, DeepSeek-V3 becomes computationally expensive with increasing prediction depth, scaling poorly for longer prediction horizons.

Our proposed method addresses these shortcomings by tightly coupling future-token predictions through a carefully structured bottleneck. This ensures the hidden state encapsulates richer predictive information without unnecessary computational overhead, highlighting the efficiency and effectiveness of JTP for representation enrichment relative to prior approaches.

\subsection{Concrete Example of the Light Module}
\label{sec:example}

We now describe a concrete example used to validate our JTP approach. In this instantiation, the $\light$ module is implemented via a single-layer self-attention mechanism.

First, we combine the hidden state $\mathbf{h}_{t-1}$ with the embeddings of the teacher-forced tokens $x_{t+j-1}$ to produce intermediate vectors $\mathbf{h}_{t-1}^{(j)}$, defined as:
\begin{align}
\mathbf{h}_{t-1}^{(j)} = \gamma \cdot \mathbf{h}_{t-1} + \mathrm{Emb}(x_{t+j-1}), \quad j=0,1,\dots,D.
\end{align}

These intermediate vectors are then processed by a single-layer self-attention module:
\begin{align}
\light_{x_{t-1:t+j-1}}(\mathbf{h}_{t-1})
= \mathbf{h}_{t-1} + \mathrm{SelfAttn}\left(\mathbf{h}_{t-1}^{(0)}, \mathbf{h}_{t-1}^{(1)}, \dots, \mathbf{h}_{t-1}^{(j)}\right) , \quad j=0,1,\dots,D.
\end{align}
The skip connection with $\mathbf{h}_{t-1}$ ensures the resulting enriched representation retains substantial context about future tokens without solely depending on teacher-forced inputs.

For consistency, we also slightly adjust the next-token prediction objective:
\begin{align}
\mathcal{L}_{\mathrm{NTP}}(x_t \mid \mathbf{h}_{t-1}) = - \log \mathrm{head}\bigl(\mathrm{SelfAttn}(\mathbf{h}_{t-1}^{(0)})\bigr)[x_t].
\end{align}
Here, since the self-attention involves only a single vector, it effectively acts as a no-op and reduces to a simple linear projection due to the value projection step.

\section{Testing Multi-Token Predictions with Star Graphs}
\label{sec:exp-star}

\begin{wrapfigure} {r}{0.35\textwidth}
    \vspace{-0.2in}
    \centering
    \includegraphics[width=0.3\textwidth]{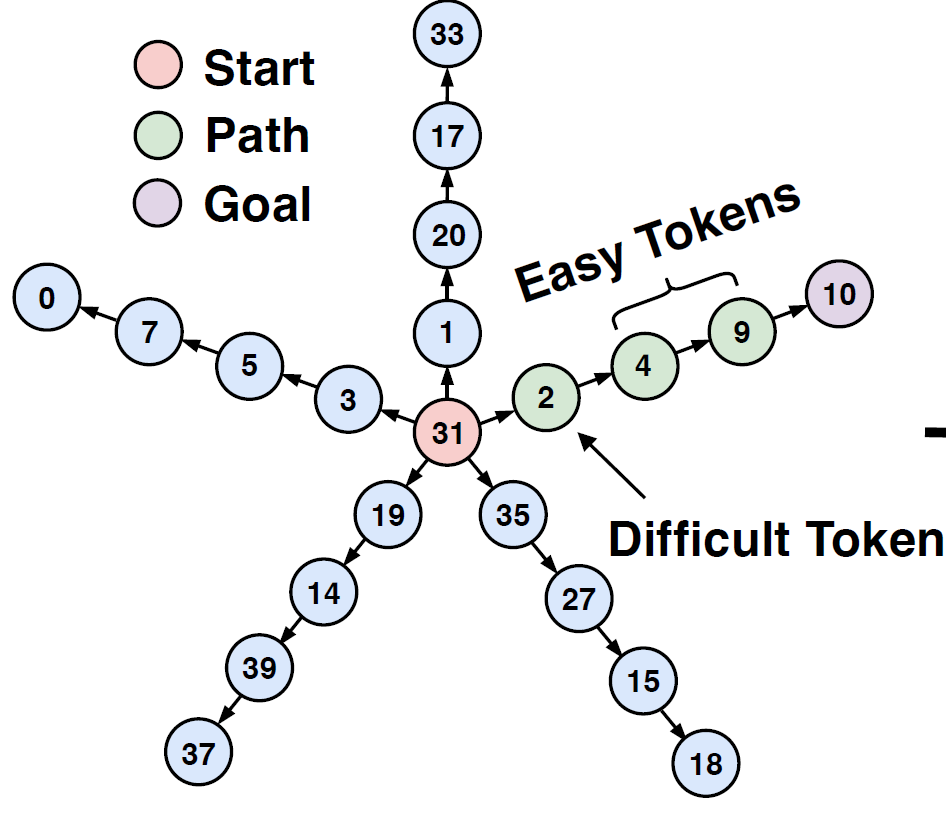}  
    \caption{Illustration of the star graph problem due to \cite{bachmann2024pitfalls}.}
    \label{fig:stargraph}
\end{wrapfigure}

We evaluate our approach using the star graph navigation task introduced by \cite{bachmann2024pitfalls}, a simple yet challenging benchmark for next-token predictors.

A star graph $G(d,l)$ (illustrated in \autoref{fig:stargraph}) consists of $d$ paths of length $l$ branching from a central node. Nodes $n_i$ are sampled uniformly from $\{1, \dots, N\}$ to construct the graph. A training example is represented as a sequence containing the edge list $\mathcal{E}$, the start and end nodes, and a path of length $l$ from start to end:
\[
\left[ \mathcal{E} \ | \ n_1, n_l\ | \ n_1, n_2, \dots, n_l \right].
\]
Despite its simplicity, modern next-token prediction models struggle to solve this task.

This problem encapsulates a fundamental challenge in planning tasks such as story writing, where maintaining coherence requires tracking both the narrative resolution and backstory while progressing through each plot point.

 \paragraph{Experimental Setting.} 
For all experiments in this section, we use a small Transformer model with $n_{\text{layers}}=6$ blocks, embedding dimension $e_{\text{dim}}=384$, $n_{\text{heads}}=8$ attention heads, and an MLP expansion factor of $e=4$. 

We implement various multi-token prediction schemes. For previous approaches \citep{gloeckle2024better,deepseekai2024v3}, we add an extra Transformer block for each depth of multi-token prediction. In contrast, our method uses only a self-attention layer without an MLP, resulting in a more lightweight module. \autoref{table:parameter} compares the parameter counts across different approaches.

All models are trained with the \textit{AdamW} optimizer \citep{loshchilov2018decoupled}, using a learning rate of $\eta = 3\cdot 10^{-4}$, $(\beta_1,\beta_2)=(0.9, 0.95)$, and a weight decay of $0.01$. We use a batch size of $1024$, with each example freshly generated for every batch.

For star graph problems, we set $N=50$, meaning node labels are sampled uniformly from ${1, \dots, 50}$. As a result, we test on graphs with at most 50 nodes.

\begin{table}[h]
    \centering
    \renewcommand{\arraystretch}{1.2}
    \begin{tabular}{l c c}
        \toprule
        Model & \# Params  \\
        \midrule
        GPT  & 10.7M \\
        JTP ($D=3$) &  11.3M  \\
        DeepSeek ($D=3$) & 13.4M  \\
        Gloeckle et al. ($D=3$) & 13.1M    \\
        \bottomrule
    \end{tabular} \quad \quad\quad  \begin{tabular}{l c c}
        \toprule
        Model & \# Params  \\
        \midrule
        GPT  & 10.7M \\
        JTP ($D=8$) &  11.3M  \\
        DeepSeek ($D=8$) & 17.8M  \\
        Gloeckle et al. ($D=8$) & 16.0M    \\
        \bottomrule
    \end{tabular}
    \caption{{\bf Parameter count comparison.} The number of parameters for different models and depths ($D=3$ and $D=8$). Our JTP model remains lightweight compared to other multi-token prediction methods. }
    \label{table:parameter}
    \label{tab:num_param}
\end{table}

\subsection{Comparison of Multi-Token Prediction Methods}

\begin{figure}
    \centering
    \begin{subfigure}{0.32\linewidth}
        \centering
        \includegraphics[width=\linewidth]{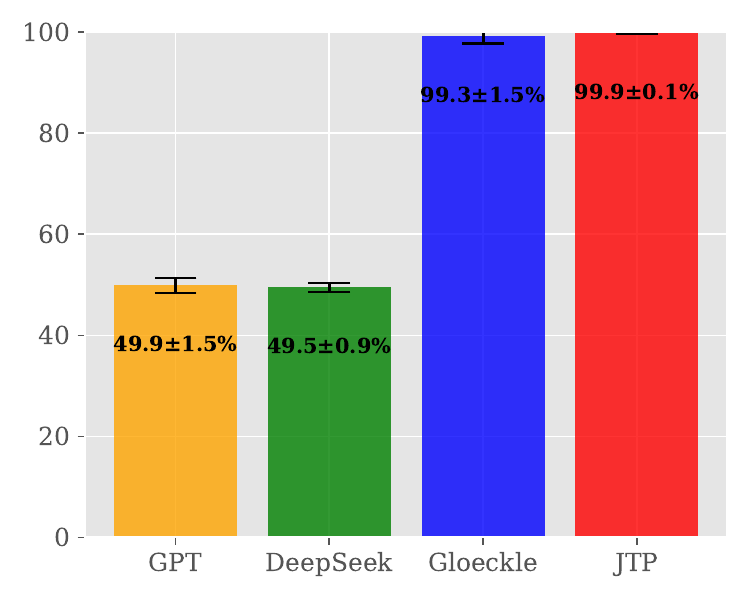}
        \includegraphics[width=\linewidth]{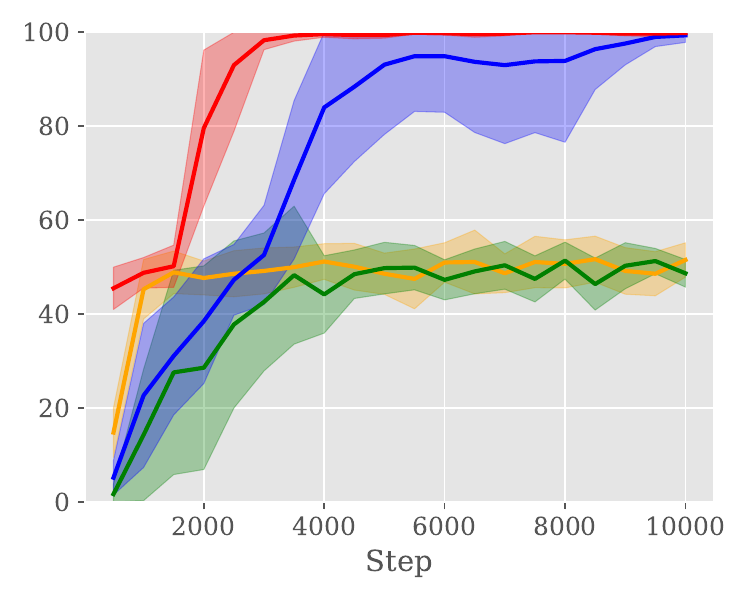}
        \caption{$G(2,5)$}
    \end{subfigure}
    \begin{subfigure}{0.32\linewidth}
        \centering
        \includegraphics[width=\linewidth]{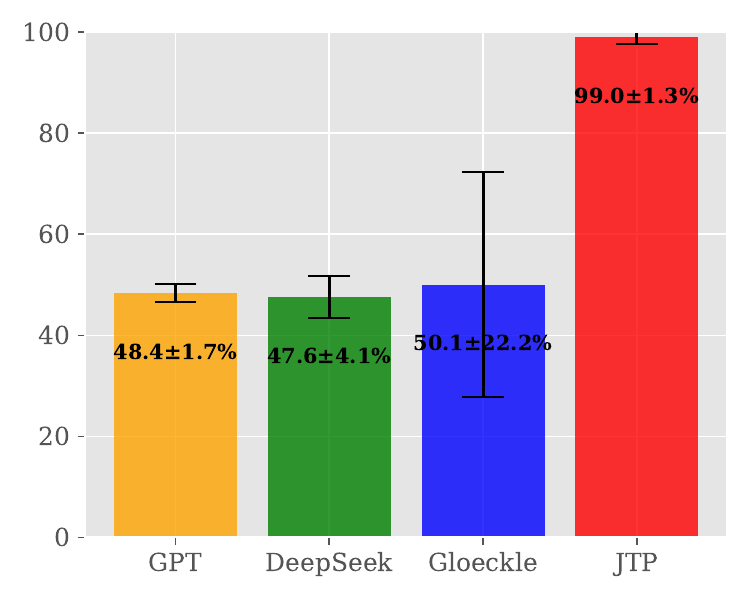}
        \includegraphics[width=\linewidth]{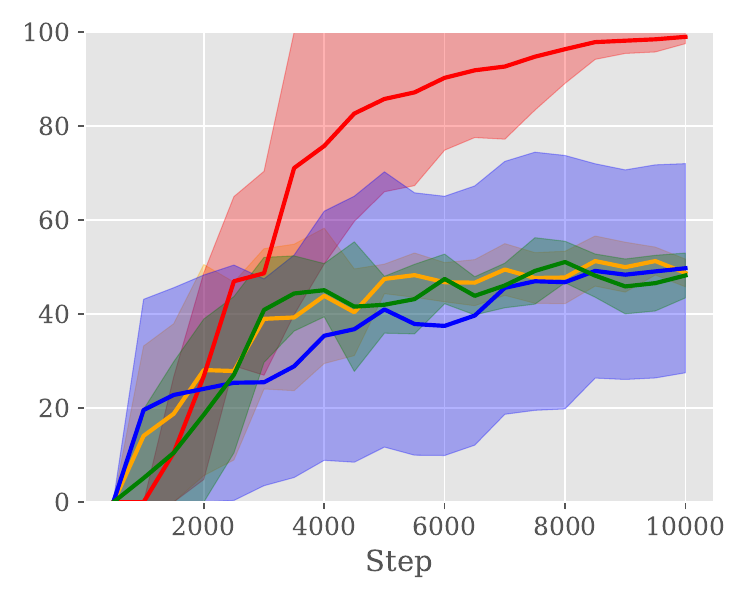}
        \caption{$G(2,10)$}
    \end{subfigure}
    \begin{subfigure}{0.32\linewidth}
        \centering
        \includegraphics[width=\linewidth]{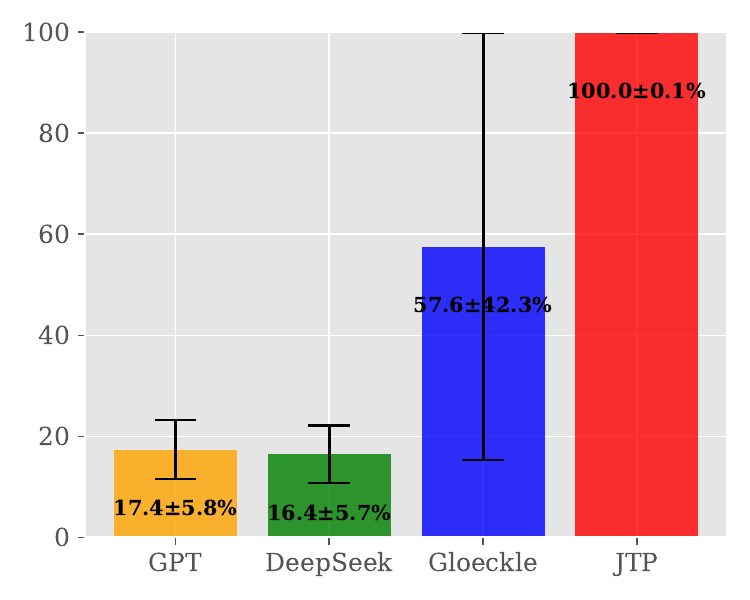}
        \includegraphics[width=\linewidth]{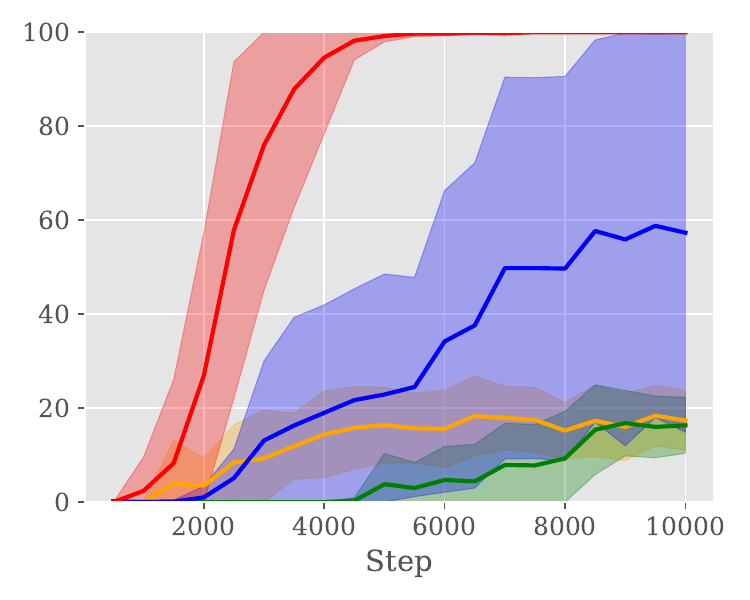}
        \caption{$G(5,5)$}
    \end{subfigure}
    \caption{{\bf Performance comparison of different methods.} Our approach consistently solves star graph tasks across different configurations, whereas prior methods struggle, especially for larger graphs.}
    \label{fig:compare}
\end{figure}

In \autoref{fig:compare}, we evaluate different multi-token prediction methods on the star graph task. For a given star graph $G(d,l)$, we set the prediction depth to $D = l - 2$, ensuring that a successful multi-token predictor can infer the critical second node by leveraging the end node of the path (see \cite{bachmann2024pitfalls} for intuition).

Our results highlight a stark contrast in performance. As expected, our approach consistently solves star graphs for $G(2,5)$, $G(2,10)$, and $G(5,5)$, while other multi-token prediction (MTP) methods struggle. The method of \cite{gloeckle2024better} successfully handles $G(2,5)$ but fails on larger graphs.

 \subsection{Small Prediction Windows}

\begin{figure} 
    \centering
    \begin{subfigure}{0.4\linewidth}
        \centering
        \includegraphics[width=\linewidth]{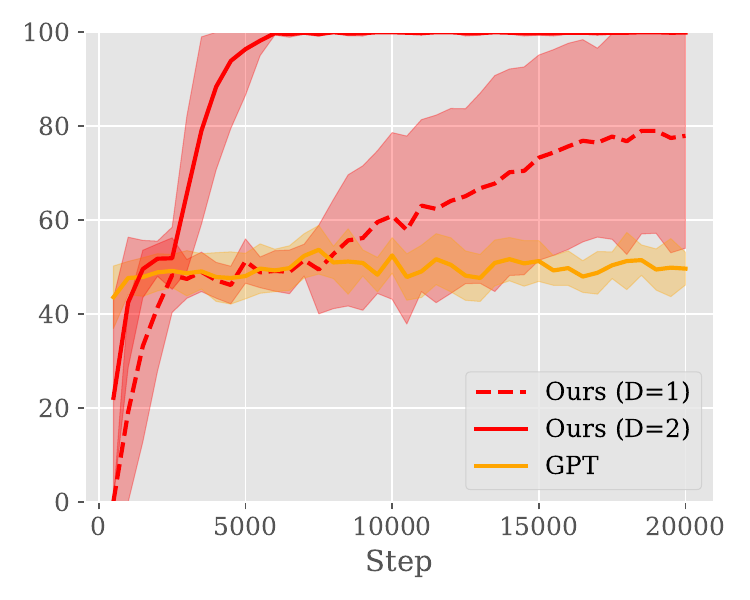}
        \caption{$G(2,5)$}
    \end{subfigure}
    \begin{subfigure}{0.4\linewidth}
        \centering
        \includegraphics[width=\linewidth]{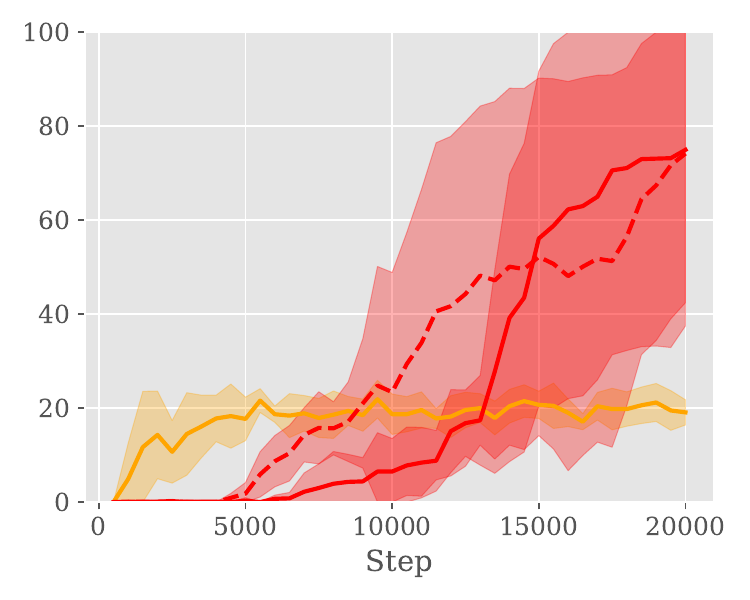}
        \caption{$G(5,5)$}
    \end{subfigure}
    
    \caption{{\bf Performance with small prediction windows.} Even at minimal depth, our JTP approach outperforms the next-token prediction baseline, demonstrating its effectiveness in shallow-depth settings.}

    \label{fig:short_graphs}
\end{figure}

In \autoref{fig:short_graphs}, we evaluate our approach with smaller prediction depths $D$. Specifically, for $G(2,5)$ and $G(5,5)$, we test $D=1,2$. 
Remarkably, even with such minimal depth, our JTP approach achieves a nontrivial improvement over the next-token prediction baseline. This highlights the effectiveness of our method, even in shallow-depth settings.

\subsection{Larger Graphs}
\begin{figure} 
    \centering
    \includegraphics[width=0.4\linewidth]{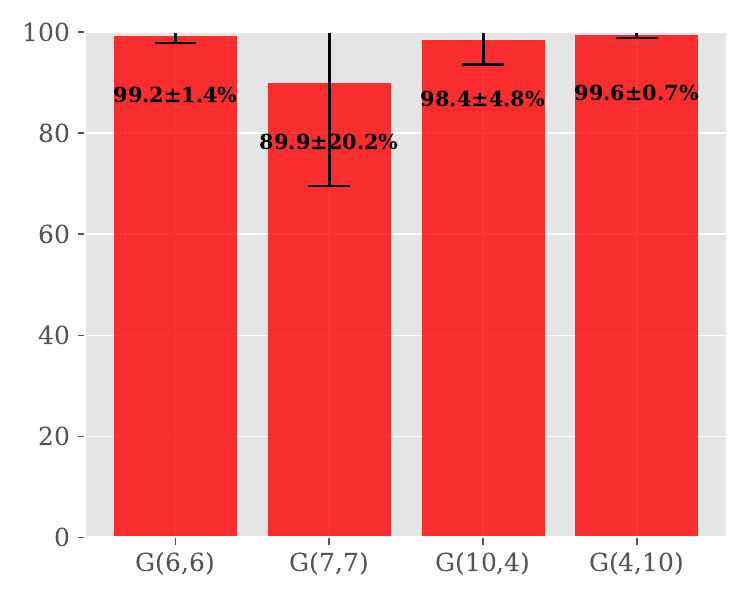}
    \caption{{\bf Performance on larger graphs.} Our approach remains effective, but for $G(7,7)$, test accuracy did not reach $100\%$ within 20,000 training steps.}
    \label{fig:large_graphs}
\end{figure}

In \autoref{fig:large_graphs}, we evaluate performance on larger graphs. Since we fix $N=50$, all tested graphs have at most 50 nodes. 

As before, for a given star graph $G(d,l)$, we set the prediction depth to $D = l - 2$, ensuring that a successful multi-token predictor can infer the critical second node using the end node of the path. Our approach remains effective across these settings, though for $G(7,7)$, it did not achieve $100\%$ test accuracy within 20,000 training steps.

\section{Theory}
\label{sec:why}
A central objective of Joint Token Prediction (JTP) is to improve the quality of learned representations used by the next-token prediction head.  We can view how JTP improves these representations from two complementary lenses: computation and completeness.

\subsection{Computation per Backward Step}

A key, and perhaps surprising, property of the JTP approach is that it uses more gradients with only negligible additional computation.   

\begin{theorem}
\label{thm:comp}
    Suppose the main Transformer has $L$ layers, and let the input sequence length be $T$.  Consider a JTP with depth $D$.  Then the following holds:
    \begin{enumerate}[label=(\roman*)]
        \item The total computation is on the order of $O\bigl(T^2 L + T D^2\bigr)$. 
        \item It provides on the order of $O(T \,D)$
        distinct gradients to the main Transformer hidden states $\mathbf{h}_{t}$ per sequence.
    \end{enumerate}
    Consequently, the flops per gradient scale on the order of $O\bigl(\tfrac{T\,L}{D} \;+\; D\bigr)$. 
\end{theorem}

\noindent\textbf{Interpretation.}  If we compare to a standard next-token-only Transformer (which typically costs $O(T^2 L)$ per forward-backward pass), the additional cost $O(T D^2)$ can be negligible for moderate $D$.  Meanwhile, the total number of gradients grows linearly in $D$.  Combining these observations, we see that for $D\ll \sqrt{T\,L}$, the added overhead remains small, yet for $D>1$ we already benefit from multiple gradients per token.  The most interesting regime for JTP is thus
\[
   D \in \bigl(1,\; c\,\sqrt{T\,L}\bigr),
\]
for some small constant $c>0$ that reflects the ratio of attention cost to head-computation cost.

\begin{proof}[Proof of  \autoref{thm:comp}]
There are two sources of computation: the Transformer blocks and the output head.  The main Transformer requires $O(T)$ computation per token implying $O(T^2)$ computation over the sequence for predictions and updates.  For each of $T$ tokens, the joint predictor head uses attention $O(D)$ times over $O(D)$ positions requiring $O(D^2)$ computation.  Critically, we note that although each extra token provides an additional gradient for the main Transformer hidden states, these gradients require \emph{no} extra computation since the activations in the bottleneck are independent of the extra tokens to be predicted.  Multiplying these together, we get $O(n D^2)$ overall computation for joint token prediction on the sequence.  Adding these two sources, we get $O(T^2L + T D^2)$ computation, establishing the first claim.

The second claim comes from noting that there are $T$ tokens each of which collects $D$ extra gradients from the joint prediction head.  Multiplying these together, we get $O(TD)$ gradients establishing the second claim.
\end{proof}

\subsection{Importance of Joint Distribution}

A key reason JTP can enrich representations is that it directly models the \emph{joint} distribution of the next $D+1$ tokens (rather than marginal distributions). By applying a teacher-forcing strategy with a representation bottleneck, the hidden state  $\mathbf{h}_t$ must contain sufficient information to generate $(x_{t+1}, x_{t+2}, \ldots, x_{t+D+1})$ in a coordinated way. 

In effect, this means $\mathbf{h}_{t}$ is forced to store all the relevant ``multi-step'' information needed to predict each of those $D+1$ tokens jointly.  This can be viewed as creating a short-horizon ``belief state'' in the sense of \cite{hu2024learning}, albeit in a much more lightweight manner.  By contrast, methods that only predict marginals (such as \cite{gloeckle2024better}) or rely on repeated re-encoding (such as \citep{deepseekai2024v3}) can circumvent this bottleneck, weakening the hidden representation.

\section{Sanity Check: Language Modeling Experiments}
\label{sec:exp-llm}

As a sanity check, we conduct preliminary language modeling experiments, summarized in \autoref{tab:language modeling}. We use a GPT-2 architecture with 162M total parameters, trained on the 5B tokens of the FineWeb dataset \citep{penedo2024the}. Our implementation is based on the \texttt{modded-nanogpt} codebase of \cite{modded_nanogpt_2024}, using the version from 11/10/24. The only modification is replacing the Muon optimizer with AdamW, using a learning rate of 0.001 for the Transformer blocks.

Importantly, we do not optimize the architecture for language modeling. Instead, we directly apply the architecture that performed well on the star graph task, using this experiment purely as a sanity check. Thus, further investigation is necessary to determine how to best adapt JTP for general language modeling.

As shown in \autoref{tab:language modeling}, JTP introduces a slight increase in next-token loss but demonstrates predictive capability in multi-token prediction. This suggests potential for further refinement in adapting JTP to language modeling.

\begin{table}[H]
    \centering
    \renewcommand{\arraystretch}{1.2}
    \begin{tabular}{l c c}
        \toprule
        Model & Final Next-Token Loss & Final Multi-Token Loss \\
        \midrule
        GPT  & 3.062 &  -- \\
        JTP ($D=1$) & 3.100 & 3.742 \\
        JTP ($D=2$) & 3.110 & 3.978 \\
        JTP ($D=4$) & 3.119 & 4.198 \\
        \bottomrule
    \end{tabular}
    \caption{{\bf Language modeling results.} JTP incurs a slight increase in next-token loss but gains predictive ability in multi-token prediction, warranting further exploration.}
    \label{tab:language modeling}
\end{table}
 
\section{Conclusion}

The next-token prediction has become a ubiquitous objective for training large language models. Recent works have attempted to build on this success by adding multi-token prediction as an auxiliary objective, with the goal of further improving the model's representation quality and training efficiency without changing the underlying architecture.  In this work, we introduced joint multi-token prediction, which uses teacher forcing through a representational bottleneck to learn a short-horizon belief state.  We found that this method improves over other multi-token prediction approaches on the Star Graph task and in a conceptual analysis.  We hope that this preliminary work will encourage more detailed study of how different multi-token prediction objectives impact representations and training dynamics.  

\bibliography{ref} 
\bibliographystyle{plainnat}
\appendix

\end{document}